\documentclass[final,3p,twocolumn]{elsarticle}

\usepackage{microtype}
\usepackage{graphicx}
\usepackage{subfigure}
\usepackage{booktabs} 

\usepackage{amssymb}

 \usepackage{lineno}

\usepackage{microtype}
\usepackage{graphicx}
\usepackage{subfigure}
\usepackage{booktabs} 

\usepackage{amsmath} 
\usepackage{mathtools}

\usepackage[hyphens]{url}
\usepackage[hidelinks]{hyperref}
\hypersetup{breaklinks=true}

\usepackage{times}
\usepackage{tikz}
\usetikzlibrary{fit,positioning,arrows.meta}
\tikzset{neuron/.style={shape=circle, minimum size=1.25cm, 
  inner sep=0, draw, font=\small}, io/.style={neuron, fill=gray!20}}

\usetikzlibrary{decorations.pathreplacing}
\usetikzlibrary{decorations.markings}

\usepackage{amssymb,amsmath}
\usepackage{helvet}
\usepackage{courier}
\usepackage{siunitx}
\usepackage{graphicx}
\usepackage{amsmath}
\usepackage{todonotes}
\usepackage[normalem]{ulem} 

\newif\ifextended
\extendedtrue
\newif\ifUSEMORE
\USEMOREtrue

\usepackage{amsmath,amsfonts,bm}

\def\eqref#1{equation~\ref{#1}}

\def\1{\bm{1}}

\DeclareMathAlphabet{\mathsfit}{\encodingdefault}{\sfdefault}{m}{sl}
\SetMathAlphabet{\mathsfit}{bold}{\encodingdefault}{\sfdefault}{bx}{n}

\DeclareMathOperator{\Proj}{proj}
\DeclareMathOperator{\prox}{prox}
\DeclareMathOperator{\st}{s.t.}

\graphicspath{{./Figures/}}

\journal{Journal of Computational Physics}

\begin{document}

\begin{frontmatter}



\title{Using Deep Learning to Extend the Range of 
       \\Air Pollution Monitoring and Forecasting}

\author[harv]{Philipp H\"ahnel}
\author[ibm]{Jakub Mare\v{c}ek}
\ead{jakub@marecek.cz}
\author[ibm]{Julien Monteil}
\author[ibm]{Fearghal O'Donncha\corref{cor1}}
\ead{feardonn@ie.ibm.com}

\address[harv]{Harvard T.H. Chan School of Public Health, Boston, Massachusetts, United States}
\address[ibm]{IBM Research, Dublin, Ireland}
\cortext[cor1]{
Corresponding author \newline}

\begin{abstract}
Across numerous applications, forecasting 
relies on numerical solvers for partial differential equations (PDEs).
Although the use  
of deep-learning techniques has been proposed, 
actual applications have been restricted by the fact the 
training data are obtained using traditional PDE solvers.
Thereby, the uses of deep-learning techniques were limited to domains, where
the PDE solver was applicable. 

We demonstrate a deep-learning framework for air-pollution monitoring and forecasting that provides
the ability to train across different model domains, as well as a reduction in the run-time by two orders of magnitude. 
It presents a first-of-a-kind implementation that combines deep-learning and domain-decomposition techniques 
to allow model deployments extend beyond the domain(s) on which the it has been trained. 
\end{abstract}


\end{frontmatter}

\section{Introduction}
Detrimental effects of air pollution on human health are long-studied. The WHO attributes 
3.8 million deaths per annum to air pollution globally \citep{WHO2018}. In many cities across 
the developed world, vehicle emissions are the dominant source of air pollutants \citep{zhang2013air}, 
contributing around 70\% of emissions of nitrogen oxides ($\mathrm{NO_x}$) \citep{londonreport}, 
and up to 50\% of particulate-matter pollution~\citep{oecd_airpoll_2014}.

Quantification, evaluation, and mitigation of these effects require systems to estimate contribution 
of traffic volumes to air pollution. Traditionally, this is done by combining estimates 
(either observed or modelled) of traffic volumes, or rather the associated pollution-generation estimates, 
with weather observations in dispersion models that resolves a set of partial differential equations (PDE) 
to compute the desired pollution distributions. This approach is limited in three ways: the PDE model 
is computationally expensive, requires considerable domain expertise, and is cumbersome to parameterise 
for further geographical locations. The computational expense restricts the meshes that can be resolved, 
in terms of the spatial extent, spatial resolution, and ultimately, the number of discrete sources of pollution.

An alternative approach leverages the capabilities of deep learning (DL) to develop rapid solvers that 
can scale to any domain size. 
As an example of the state of the art in the area, \citet{Fearghal2018} recently reported a factor 
of 5,000 computational speed-up compared to that of a leading PDE solver.
Considering that the PDE solver is used to generate the outputs to train the DL model on, however, the model is 
still limited to the domain that the PDE solver can be run on.
We integrate DL together with techniques from PDE-based domain decomposition 
to present an approach that learns the pollution dispersion at independent, neighbouring meshes and merges 
what has been learned into a single unified model for the region. More specifically, 
 a PDE model for air pollution is used to  generate sufficiently large volumes of data
to facilitate the use of surrogate or reduced-order models  \citep{benner2015survey} using deep artificial 
neural networks \citep{goodfellow2016deep}. The model is deployed independently for a series of meshes with 
each representing a subset of a geographical domain and the DL model trained on the outputs of these meshes 
and merged in a recurrent neural networks (RNN) \citep{rumelhart1986learning} type implementation to provide a 
single DL model of the entire region. The surrogate model serves as a computationally lightweight representation
of the PDE-based model. 

In our approach, the DL was trained on small domains and then applied to larger domains, with consistency constraints 
ensuring that the solutions are physically meaningful even at the boundaries of the domains. Our contributions are as follows:
\begin{itemize}
\item definition of the consistency constraints, 
wherein the output for one mesh is used to constrain 
the output for another mesh. 
\item methods for applying the consistency constraints within the training of a DL,
which allows for an increase in the extent of the spatial domain by concatenating the 
outputs of several PDE-based models and considering conditions at neighbouring mesh interface.
\item a numerical study of the approach on a pollution-forecasting problem, 
wherein we 
show the test mean absolute error (MAE) against PDE model 
and sensor data.
\end{itemize}

\section{Related Work}

A long-standing challenge in applied mathematics is the boundary-value problem, 
which consists in imposing boundary conditions at the defined internal or external boundaries of the region that is governed 
by a PDE. Such boundary conditions are additional constraints that usually come from field 
measurements, change of topology, or external models. Ensuring that the PDEs are solved, while embedding the boundary conditions, is 
the challenge of many practical engineering applications. Examples include resolution of the Saint-Venant~\citep{barre} 
and Lighthill--Whitham--Richards (LWR)~\citep{lighthill1955kinematic} equations, 
for water and transport network problems, respectively.
In atmospheric pollution modelling, one considers the advection-diffusion 
equation~\cite[see equation 2.2]{stockie2011mathematics} and the stochastic-coagulation 
equation~\cite{norris1999smoluchowski}, while financial modelling often considers Black-Scholes models 
and its numerous (stochastic) variations~\citep{ronnie1998general}. 
Throughout, there are different types of boundary problems 
depending on whether the function, derivative, or variable itself is known at the boundary. 
Among the numerical methods to solve this boundary-value problem for PDEs, prominent examples are the 
Godunov's scheme~\citep{godunov1959difference},
 which involves solving a Riemann problem~\citep{leveque1990conservative} at each defined 
 cell interface, 
the Lax-Friedrichs method~\citep{lax1957hyperbolic}, which relies on the introduction of 
a viscosity term, and the Galerkin method~\citep{donea1984taylor}. 
The interested reader can refer to e.g.~\cite{leveque1990conservative}
for a discussion on the topic. 

In practice, such discontinuities may be either inherent to the physics itself
 (proper boundary conditions, different models due to different flux functions, e.g.,
   change from a motorway traffic network to a urban traffic network, different data 
   sources for different regions in space) 
 or artificial (software limitations, limited run-time, 
	 different entities providing different models). 
To our knowledge, our proposed approach is the first to address the discontinuities caused by the 
(arbitrary or not) discretisation in space of the grid where the forecasting is done via surrogate 
DL model(s) to the underlying PDE model(s). 
 
 In the problem presented in this paper, two classes of boundary conditions are considered:  
 first, external boundary conditions that describe influx of pollutants to the domain (e.g., traffic volumes, 
wind conditions), and second, internal boundary conditions that can be enforced to ensure consistency between 
neighbouring domains. The latter can be considered a variant of the additive Schwarz method widely applied 
in the solution of partial differential equations. The additive Schwarz method provides an approximate 
solution for a boundary value problem by splitting it into boundary value problems on smaller domains and 
adding the results \citep{ragnoli2019localised}. These domain-decomposition techniques are widely studied in parallel-computing applications 
and consist of solving subproblems on various subdomains while enforcing suitable consistency constraints
 between adjacent subdomains, until the local problems converge to an approximation of the true solution \citep{odonncha2019allscale}.
 It proceeds by  splitting the global domain into a set of smaller overlapping or non-overlapping subdomains. 
 In each step, an iterative method resolves the partial differential equations restricted to individual subdomains 
 and then coordinates the solution between adjacent subdomains. 
 In our study, we consider an approach that trains a DL model on a set of subdomains with an iterative reduction 
 used to enforce a physically meaningful relationship between adjacent subdomains. 
 A large number of domain-decomposition approaches exist and we refer to the books by \citet{quarteroni_domain_1999} 
 and \citet{smith_domain_2004} for an extensive introduction. 

		 Our methods draw on 
		 domain-decomposition implementations and recent work within applying DL to PDE-based models.  
		 A large body of literature exist on the use of neural networks with PDE-based models \citep{712178,870037,lee1990neural,1528518,1716337,5179018,
		 6658964,rudd2013solving}, 
		 across a variety of applications such as fluid modelling \cite{brunton2019machine}, combustion modelling \cite{Wang2019}, and the geosciences \citep{lary2004using,loyola2006applications}. More recently, the use of DL as surrogate for expensive physics-based models has been investigated  \citep{tompson2017,Fearghal2018,wiewel2018latent,Hesthaven2018,Wang2019,Guo2019}.  In these latter studies, a solver for PDEs has been used to obtain hundreds of thousands
		 of outputs corresponding to hundreds of thousands of inputs.
		 The DL has then been used as means of non-linear regression between the inputs and
		 outputs.
		 In particular, \citet{tompson2017} considered a convolutional neural 
		 network (CNN), 
		 while \citet{wiewel2018latent} considered 
		 long short-term memory (LSTM) units within a recurrent neural network (RNN).
		 In a more abstract setting, \citet{sirignano2017dgm} have explored the use of 
		 mesh-free DL in what they call deep Galerkin methods. 
		 Throughout, the applications of DL have been limited in scale to the domains
		 that had been tractable for the traditional solver for PDEs.

		 A wide-variety of recent applications of deep-learning in numerical analysis include  non-intrusive reduced basis \cite{Hesthaven2018} and related \cite{Wang2019,Guo2019} methods for the construction of reduced-order models, learning PDEs from data \cite{long2019pde,qin2019data},
		 and 
		 the detection of troubled cells in two-dimensional unstructured grids \cite{Ray2018,Ray2019}.
		 
		 A further stream of related work has been started by \citet{NIPS2018_7892}, who presented a novel approach to approximate the 
		 discrete series of layers between the input and output state by acting on the derivative of 
		 the hidden units. At each stage, the output of the network is computed using a black-box 
		 differential equation solver which evaluates the hidden unit dynamics to determine the
		 solution with the desired accuracy.  In effect, the parameters of the hidden unit dynamics 
		 are defined as a continuous function, which may provide greater memory efficiency and
		 balancing of model cost against problem complexity.  The approach aims to achieve comparable
		 performance to existing state-of-the-art with far fewer parameters, and suggests potential 
		 advantages for time series modelling.
For follow-up work in this stream, see \cite{grathwohl2018scalable,gulgec2019fd}.
		 On a similar note \citet{han2018solving} investigated approaches to solve PDEs using deep learning 
		 gradient-based approaches. By reformulating the PDEs  as backward stochastic differential equations 
		 the unknown is solved for using a gradient-descent approach based on reinforcement learning.
		 
		 Finally, there is a long history of the use of 
		 machine learning in pollution monitoring \citep[e.g.]{Thomas2007,DONG2009,qi2017deep,zhu2018machine}. 
		 Recently, \citet{zhu2018machine} considered a coarse (0.25 degree resolution) 
		 grid of mainland China, with more than two years of air quality measurement and 
		 meteorological data, without any further insights, such as pollution sources, 
		 surface roughness, the reaction model, the multi-resolution aspects, or similar.
		 \citet{qi2017deep}, considered a joint model for feature extraction, interpolation, and prediction 
		 while employing the information pertaining to the unlabelled 
		 spatio-temporal data to improve the performance of the predictions. 
		 These approaches use the measurement data without regard to the physics,
		 which limits their performance, given the sparsity and costs of presently available sensors.


		 \section{Our Approach}

We present the first attempt to apply a domain decomposition to training of a surrogate model of a
 partial differential equation (PDE). 
At a high level, our approach consists of training a deep-learning model 
for each subdomain, while providing a method to ensure consistency across neighbouring domains. 
By enabling communication between subdomains via constraints, predictions for one subdomain can benefit from information outside of the subdomain.
This makes it possible to scale
to the whole domain such that the accuracy of the predictions and its ability to generalise is increased, compared to models trained on the individual subdomains without the consistency constraints. 

Let us consider an index-set $\mathcal{M}$ of meshes $M_m$, $m \in \mathcal{M}$, with sets of $n_m$ mesh points $P(M_m)$.
The output of each PDE-based simulation on such a mesh consists of values in $\mathbb{R}^{d_m}$ at each point of $P(M_m)$.
Often, a small sub-set of $n_m^{(r)}$ of such points is of particular interest, 
which we call receptors $R(M_m)$;
the remainder of the points represents hidden points $H(M_m)$. 
The receptors and hidden points thus partition the mesh points $P(M_m) = H(M_m) \cup R(M_m)$, with $n_m = n^{(h)}_m + n^{(r)}_m$. 
Further, let us consider the index-set $\mathcal{B} \subseteq \mathcal{M} \times \mathcal{M}$ of boundaries
$B_{mn}$ of meshes.
Such a possibly infinite boundary $B_{mn} \subseteq P(M_m) \times P(M_n)$ links pairs of points from the two meshes. To each boundary $B_{mn}$ we associate a constant $\epsilon_{mn}$ that reflects the importance of this boundary.
Further, for each mesh $M_m$ we have an ordered set of simulations indexed with time $t \in \mathbb{Z}$, 
where each simulation is defined  
by the inputs $x^{(m)}_t \in X^{(m)}_t$ and a set of outputs $y^{(m)}_t \in (\mathbb{R}^{d_m})^{\times n_m}$. 
Often, one wishes to consider $y^{(m)}_t$ being part of $x^{(m)}_{t+k}$ for some $k > 0$, in a recurrent fashion.

Our aim is to minimise residuals subject to consistency constraints, and thus exchange information between 
neighbouring domains and ensure physical ``sanity'' of the results, i.e.,
\begin{align}
  \label{infinite} 
     r^* = &
    \min_{f} \sum_{t} \sum_{m \in \mathcal{M}} \left\| \Proj_{R(M_m)}\! \left( y^{(m)}_t -  f^{(m)}\!\big(x^{(m)}_t\big) \right)\right\|  
    \\    &
    \st \;\forall t \; \forall (m, n) \in \mathcal{B} \; \forall (p_1, p_2) \in B_{mn} : \notag \\
    &  \prox\!\left( \Proj_{\{p_1\}}\! f^{(m)}\big(x^{(m)}_t\big)\,, \; \Proj_{\{p_2\}}\! f^{(n)}\big(x^{(n)}_t\big) \right) \le \epsilon_{mn} \,,
 \notag
\end{align}
where $\Proj_{Q} : (\mathbb{R}^{d_m})^{\times n_m} \to (\mathbb{R}^{d_m})^{\times |Q|}$ is a projection operator that projects the array of outputs at all points onto the outputs at a subset of points $Q \subset P(M_m)$, 
$\prox$ is a proximity operator, 
the decision variable defines the mapping $f = \{f^{(m)}\}^{m \in \mathcal{M}}$,   whereby  $f^{(m)}\!\big(x^{(m)}_t\big)$ represents the output of a non-linear mapping between inputs and PDE-based simulation outputs at the points of the mesh,
$f^{(m)} : X^{(m)}_t \to (\mathbb{R}^{d_m})^{\times n_m}$, on each independent mesh $M_m$, which can be seen
as a non-linear regression, and $\epsilon_{mn}$ is a constant specific to $(m, n) \in \mathcal{B}$.
Thinking about a system based on a PDE, the projection operator onto $R(M_m)$ can be thought of as selecting the receptors, which are positions at which the solution to the PDE is evaluated. In principle, the set of mesh points can also contain points for which no estimates from the ground-truth model are generated. 

Notice that 
$f^{(m)} : X^{(m)}_t \to (\mathbb{R}^{d_m})^{\times n_m}$
should be seen as a non-linear regression;
we provide examples of $f^{(m)}$ in the following sections.
The requirement of physical ``sanity'' is usually a statement about smoothness of the values of the mapping $f^{(m)}$ across the boundaries of two different meshes and represents the fact that processes in one mesh impact direct neighbours.
To be able to compare those values, we require that the dimensions are the same, that is $\forall m, n \in \mathcal{M}: \; d_m = d_n \equiv d $.
For example, for $\prox$ being the norm of a difference of the arguments, ``smooth'' at a point at the boundary of two meshes means that the values predicted within the two meshes at that point are numerically close to each other. 
Also adding the norm of the difference of their gradients to that makes it a statement about the closeness of their first derivatives too. Technically, ``smoothness'' is a statement about all their higher derivatives as well, however, we will only concern ourselves with their values, or zeroth order of derivatives, for now. 
Notice though that generically this is an infinitely large problem.
%

The constrained optimisation problem \eqref{infinite} can be solved by Lagrangian relaxation techniques 
\citep{fisher1981lagrangian}, wherein for Lagrange multipliers $\lambda :=  \{\lambda^{(m)}_t\}^{m \in \mathcal{M}}_{t \in \mathbb{Z}}$
 we have an unconstrained optimisation problem, as suggested in equation~\ref{infinite-dimensional} in Figure~\ref{fig:infinite-dimensional}
 
 \begin{figure*}[tb]
 \begin{align}
\label{infinite-dimensional}
    r^* = &
    \inf_{f,\, \lambda}  \sum_{t} \Bigg(  \sum_{m \in \mathcal{M}} \left\| \Proj_{R(M_m)}\! \left( y^{(m)}_t - f^{(m)}(x^{(m)}_t) \right) \right\| \\
    & + 
    \sum_{(m, n) \in \mathcal{B}} \sum_{(p_1, p_2) \in B_{mn}}  \notag 
    \lambda^{(m)}_t \prox\!\left( \Proj_{\{p_1\}}\! f^{(m)}\big(x^{(m)}_t\big)\,, 
        \Proj_{\{p_2\}}\! f^{(n)}\big(x^{(n)}_t\big) \right) \Bigg)\, \notag
\end{align}
\caption{The Lagrangian relaxation.}
\label{fig:infinite-dimensional}
\end{figure*}

Under mild conditions \citep[Proposition 3.1.1]{bertsekas1997nonlinear},
 there exist $\lambda_t^{(m)}$, $t \in \mathbb{Z}$, such that the infimum over $f^{(m)}$ coincides with $r^*$, for each $m \in \mathcal{M}$.
Clearly, if at least some of the boundaries $B_{mn}$ are infinite, then 
the optimisation problem is infinite-dimensional.

Next, one can borrow techniques from iterative solution schemes 
in the numerical analysis domain. 
Notice that the first term in \eqref{infinite-dimensional}
 is finite-dimensional and 
separable across the meshes.
For each mesh $M_m$, $m \in \mathcal{M}$, the above can be computed independently. 
Further, one can sub-sample the boundaries
to obtain a consistent estimator.
Subsequently, one could solve the finite-dimensional projections of \eqref{infinite-dimensional}, 
wherein each new solution will increase the dimension of $\lambda^{(m)}_t$.
While this is feasible in theory, the inclusion of non-separable terms with $\lambda^{(m)}_t$ would still render the solver less than practical.

Instead, we propose an iterative scheme, which is restricted to separable approximations.
Let us imagine that during iteration $k$ and at time $t$, for a pair of points $(p_1, p_2) \in B_{mn}$ 
on the boundary indexed with $(m, n) \in \mathcal{B}$, we obtain values from the trained model at those points in the respective mesh,
$\mathbb{R}^{d} \ni f^{(m)}_{p_1, t} = \Proj_{\{p_1\}}f^{(m)}\big(x^{(m)}_t\big)$ and $\mathbb{R}^{d} \ni f^{(n)}_{p_2, t} = \Proj_{\{p_2\}}f^{(n)}\big(x^{(n)}_t\big)$. 
While the two points $p_1, p_2$ lie in two different meshes, we would like the model outputs at those points to eventually coincide for high enough $k$. For that, we iteratively construct vectors $\underline \chi_{p_1, p_2}^{(k+1)}$ and $\overline \chi_{p_1, p_2}^{(k+1)} \in \mathbb{R}^d$, which serve as lower and upper bounds on the values obtained from the training of $f^{(m)}$ at the $k^{\rm th}$ iteration. Those vectors can be updated through a variety of methods. A na\"ive example includes extracting some bounds based on the upper and lower limits of neighbouring meshes boundary values. 
In order to obtain convergence properties in a non-convex setting, we use an asymptotically vanishing shift term to adjust the interval according to the newly trained data, and a gradient term, according to

{\small
\begin{align}
    {\underline \chi_{p_1, p_2}^{(k+1)}}_{i} = 
        {\underline \chi_{p_1, p_2}^{(k)}}_{i}
        & + \frac{\kappa}{\sqrt{k} + \zeta} \left({\rm min}\!\left( {f^{(m)}_{p_1, t}}_i, \, {f^{(n)}_{p_2, t}}_i \right) - {\underline \chi_{p_1, p_2}^{(k)}}_{i}\right) \notag \\
        & + 
        \frac{\kappa\sqrt{k}}{\sqrt{k} + \zeta} \left({\underline \chi_{p_1, p_2}^{(k)}}_{i} - {\underline \chi_{p_1, p_2}^{(k-1)}}_{i} \right)
    \,, \nonumber\\
    {\overline \chi_{p_1, p_2}^{(k+1)}}_{i} = 
        {\overline \chi_{p_1, p_2}^{(k)}}_{i}
        & + \frac{\kappa}{\sqrt{k} + \zeta} \left({\rm max}\!\left( {f^{(m)}_{p_1, t}}_i, \, {f^{(n)}_{p_2, t}}_i \right) - {\overline \chi_{p_1, p_2}^{(k)}}_{i}\right)  \notag \\
        & +
         \frac{\kappa\sqrt{k}}{\sqrt{k} + \zeta} \left({\overline \chi_{p_1, p_2}^{(k)}}_{i} - {\overline \chi_{p_1, p_2}^{(k-1)}}_{i} \right)
    \,.
\label{eq:updates}
\end{align}
}

The free parameters $\kappa$ and $\zeta$ are tunable 
and resemble learning rates. In principle, they could be chosen dynamically, specific to each boundary $(m,n)$. Choosing them to be constants based on a greedy search across a limited parameter space eases the computational efforts.

For the first iterations, the boundary values are initialised using the minimum and maximum of the labels, respectively. 
Subsequently, we can form univariate (box, interval) constraints, restricting the corresponding elements of both $f^{(m)}$ at $p_1$ and $f^{(n)}$ at $p_2$ of the next iteration to the interval 
$\big({\underline \chi_{p_1,p_2}}_i, {\overline \chi_{p_1,p_2}}_i\big)$. 
Notice that replacing $\lambda^{(m)}_t$ with a constant $\lambda$ provides an upper bound on $r^*$, which is much easier to solve, computationally.

In the scheme, we consider a finite-dimensional projection of \eqref{infinite-dimensional}.
For each $(m, n) \in \mathcal{B}$ we consider a finite sample $\hat B_{mn} \subset B_{mn}$ of pairs of points, for which we obtain
{\small 
\begin{align}
\label{eq:ROM}
    r^* & = 
    \min_{f, \lambda} \sum_{t} \Bigg( \sum_{m \in \mathcal{M}} \left\| \Proj_{R(M_m)}\left( y^{(m)}_t - f^{(m)}\big(x^{(m)}_t\big) \right) \right\|
\\ 
    & \quad + \sum_{\substack{(m, n) \in \mathcal{B} \\
    (p_1, p_2) \in \hat B_{mn}}} \sum_{\substack{l \in (m, p_1)\\p \in  (n, p_2)}} \notag \\
    &
    \epsilon_{mn}
    \lambda \left\| 
    \max\!\big(0, \underline \chi_{p_1,p_2} - f^{(l)}_{p, t}\big)  
 +     \max\!\big(0, f^{(l)}_{p, t} - \overline \chi_{p_1,p_2}\big) \right\|_1\! \Bigg)\,, 
 \notag
\end{align}}
\noindent
where we consider the function $\max: \mathbb{R}\times\mathbb{R}^d \to \mathbb{R}^d$ to operate element-wise. 
Further, when we consider $\lambda$ as a hyper-parameter, we obtain an  optimisation problem separable in $m \in \mathcal{M}$, which in the limit of $|\hat B_{mn}| \to  |B_{mn}|$ provides an over-approximation for any $\lambda$.

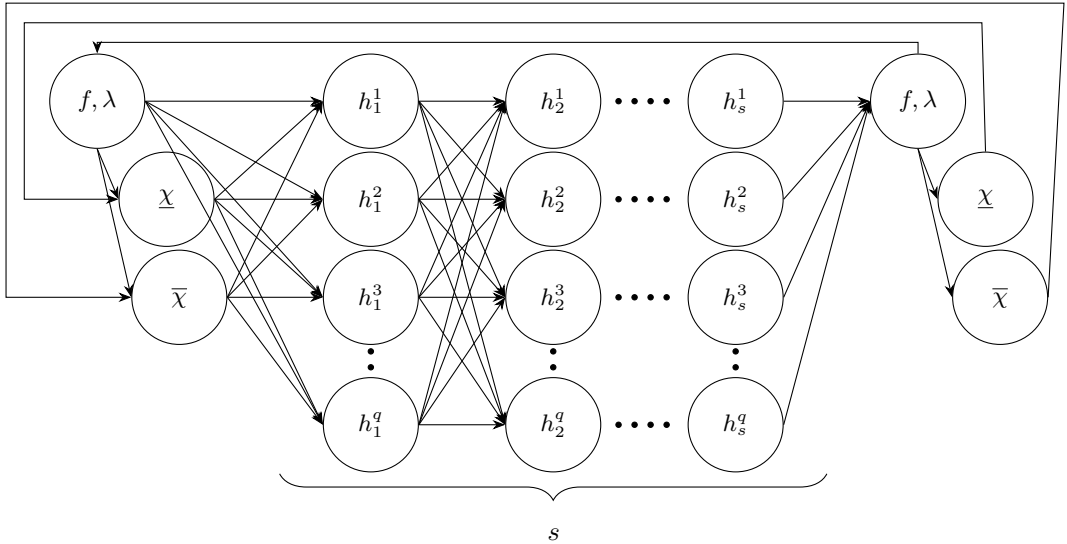
\begin{figure*}[ht!]
\centering
\begin{tikzpicture}[x=2.4cm, y=1.3cm, >=Stealth,
decoration={markings,
  mark=between positions 0.2 and 0.9 step 6pt
  with { \draw [fill] (0,0) circle [radius=1.0pt];}}]

      \node [neuron] at (0.5, 1-1) (h-1-1){$f, \lambda$};
      \node [neuron] at (0.88, 1-2) (h-2-1){$\underline \chi$};
      \node [neuron] at (0.95, 1-3) (h-3-1){$\overline \chi$};
      
      \draw [->] (h-1-1.south) -- (h-2-1.west);
      \draw [->] (h-1-1.south) -- (h-3-1.west);

\draw [decorate,decoration={brace,mirror,amplitude=10pt},xshift=0.0cm,yshift=0pt]
      (1.5,-3.8) -- (4.5,-3.8) node [midway,right,xshift=-0.2cm,yshift=-0.8cm] {$s$};

\node [neuron] at (2, 1-1) (h-1-2){$h_{1}^{1}$}; 
\node [neuron] at (2, 1-2) (h-2-2){$h_{1}^{2}$}; 
\node [neuron] at (2, 1-3) (h-3-2){$h_{1}^{3}$}; 
\node [neuron] at (2, 1-4.3) (h-4-2){$h_{1}^{q}$}; 

      \foreach\i in {1, 2, 3}
        \foreach \ii in {1, 2, 3, 4}
           \draw [->] (h-\i-1.east) -- (h-\ii-2.west);
      \path[postaction={decorate}] (h-3-2) to (h-4-2);

\node [neuron] at (3, 1-1) (h-1-3){$h_{2}^{1}$}; 
\node [neuron] at (3, 1-2) (h-2-3){$h_{2}^{2}$}; 
\node [neuron] at (3, 1-3) (h-3-3){$h_{2}^{3}$}; 
\node [neuron] at (3, 1-4.3) (h-4-3){$h_{2}^{q}$}; 

     \foreach\i in {1, 2, 3, 4}
        \foreach \ii in {1, 2, 3, 4}
           \draw [->] (h-\i-2.east) -- (h-\ii-3.west);
     \path[postaction={decorate}] (h-3-3) to (h-4-3);

\node [neuron] at (4, 1-1) (h-1-s){$h_{s}^{1}$}; 
\node [neuron] at (4, 1-2) (h-2-s){$h_{s}^{2}$}; 
\node [neuron] at (4, 1-3) (h-3-s){$h_{s}^{3}$}; 
\node [neuron] at (4, 1-4.3) (h-4-s){$h_{s}^{q}$}; 
\path[postaction={decorate}] (h-3-s) to (h-4-s);

\path[postaction={decorate}] (h-1-3) to (h-1-s);
\path[postaction={decorate}] (h-2-3) to (h-2-s);
\path[postaction={decorate}] (h-3-3) to (h-3-s);
\path[postaction={decorate}] (h-4-3) to (h-4-s);

      \node [neuron] at (5, 1-1) (h-1-r){$f, \lambda$};
      \node [neuron] at (5+0.37, 1-2) (h-2-r){$\underline \chi$};
      \node [neuron] at (5+0.45, 1-3) (h-3-r){$\overline \chi$};

      \draw [->] (h-1-r.south) -- (h-2-r.west);
      \draw [->] (h-1-r.south) -- (h-3-r.west);
      
      \foreach\i in {1, 2, 3, 4}
        \foreach \ii in {1}
           \draw [->] (h-\i-s.east) -- (h-\ii-r.west);
           
    \draw [->] (h-1-r.north) -- (5, 0.6) -- (0.5, 0.6) -- (h-1-1.north);
    \draw [->] (h-2-r.north) -- (5.35, 0.8) -- (0.1, 0.8)-- (0.1, 1-2)-- (h-2-1.west);
    \draw [->] (h-3-r.east) -- (5.8, 1.0) -- (0.0, 1.0)-- (0.0, 1-3)-- (h-3-1.west);


\end{tikzpicture}
\caption{A schematic illustration of our recurrent neural network, where the recursion considers the consistency constraints defined by $\underline \chi, \overline \chi$. In experiments described in Section~\ref{sec:meth}, we use $q=50$ and $s=7$.}
\label{fig:RNN}
\end{figure*}

In deep learning, this scheme should be seen as a recurrent neural network (RNN). A fundamental extension of RNN compared to traditional neural network approaches is parameter sharing across different parts of the model.
We refer to \citet{goodfellow2016deep} for an excellent overview and Figure~\ref{fig:RNN} for a schematic illustration. 
Each training iteration provides constants
$(\underline \chi_{p_1,p_2}, \overline \chi_{p_1,p_2})$, 
which are used in the consistency constraints of the subsequent iteration. 
In terms of training the RNN, it is important to notice that \eqref{eq:ROM} allows for very fast convergence rate even in many classes of non-linear maps $f$. For instance, when $f^{(m)} : X^{(m)}_t \to (\mathbb{R}^{d_m})^{\times n_m}$ is a polynomial of a fixed degree \citep{gergonne1974application}, then \eqref{eq:ROM} is strongly convex, despite the max function making it non-smooth. 
The subgradient of the max function is well understood  \citep{boyd2004convex} and readily implemented in major deep-learning frameworks. 

In numerical analysis, in general, and with respect to the multi-fidelity methods \citep{peherstorfer2018survey}, in particular, 
our approach could be seen as iterative  model-order reduction. 
The original PDEs could be seen as the full-order model (FOM) to reduce, and
\eqref{infinite} could be seen as a high-fidelity data-fit reduced-order model (ROM), albeit not a very practical one, whereas 
\eqref{eq:ROM} could then be seen as a low-fidelity data-fit ROM, which allows for rapid prediction.

In approximation theory, and learning theory that grew out of it, it is known since the work of \cite{Cybenko1989} that even a feed-forward network with three or more layers of a sufficient number of neurons with, e.g., sigmoidal activation function allows for a universal approximation of functions on a bounded interval. It is not guaranteed, however, that the approximation has any further desirable properties, such as energy conservation etc. Our consistency constraints can be used to enforce such properties.

Fundamentally, the approach can be summarised as learning the non-linear mapping between 
inputs and predictions on each independent mesh, and iterating to
ensure consistency of the solution across meshes. 
Such an approach draws on a long history of work on setting boundary conditions as consistency constraints  
in the solution of PDEs \cite{quarteroni1999domain}. 


\section{Methods}
\label{sec:meth}

To illustrate this framework, we trained the DL for city-scale pollution monitoring, utilising:
\begin{itemize}
    \item Pollution measurements and traffic data.
    \item Weather data (i.e., velocities, pressures, humidity, and temperatures in 3D).  
	\item A given discretisation of a city in multiple meshes, corresponding to multiple geographic areas with their specificities.
\end{itemize}

Our test case was based in the city of Dublin, Ireland, for which real-time streams of traffic and pollution data (from Dublin City Council), and weather data (from the Weather Company) were available to us, but which did not have any large-scale models of air pollution deployed.


\subsection{Air Pollution Forecasting}

Our goal is to estimate the traffic-induced air pollution, specifically
 levels of NO$_2$ (which is closely related to $\mathrm{NO}_{\text{x}}$ overall) and $\mathrm{PM}_{10}$, for defined receptors across the city. We selected these pollutants and as they are central to major public health concerns particularly in relation to an observed increase of lung and heart diseases in cities across the developed world, and as they are mostly generated by vehicle emissions. In addition ozone $\mathrm{O}_3$ is typically produced as a result of complex reactions involving organic compounds as well as nitrogen oxides $\mathrm{NO}_{\text{x}}$. 

Our prediction framework consisted of inputs of traffic volumes for a number of roadway links across the city, weather data, and an air-pollution dispersion model. Outputs consisted of periodic estimates of pollution levels. The typical approach in the traffic-induced air-pollution forecasting 
literature is to treat links as line sources. 
Dispersion models are, in fact, line-source models that describe the temporal and spatial evolution of vehicle emissions 
near roadways~\cite{NAGENDRA20022083}. Gaussian-plume models consider a closed-form solution to the advection-diffusion equation 
under a series of simplifying and steady-state assumptions, see from equation 2.2 in~\cite{stockie2011mathematics}. The Caline~\cite{caline}, Hiway~\cite{hiway}, and Aermod~\cite{aermod} suite of models are three examples of Gaussian plume EPA-developed models, 
while the latest releases of Caline (Caline 3 and 4) have had the widest adoption over the last decades, see e.g.~\citet{samaranayake2014real}, and Aermod is the 
most recently developed, but licensed model. There also exist more sophisticated numerical models in the literature. One can name the non-steady-state Lagrangian puff modelling system, calPuff~\cite{scire2000user} or even CFD models such as the commercial Ansys Fluent model~\cite{fluent} relying on the discretization of air flow variables as finite control volumes. Finally, the machine learning family of air dispersion models should be mentioned, e.g. from multivariate analysis~\cite{hsu1992time} to neural-network models~\cite{hossain2014predictive}.

 In this work, we choose to use a Gaussian-plume model in its popular Caline-4 implementation, due to its wide use across the years. We are aware of the complexity of some of the air dispersion models of use in the literature.
 However, we selected the Gaussian plume model and its Caline 4 implementation for the sake of simplicity and reproducibility. Dealing with more complex models within our presented framework is not an impossible task, but would be an engineering challenge requiring the handling of possibly more air quality variables and possibly deeper DL models. This falls beyond the scope of this paper which aims at providing a proof of concept of our newly presented methodology.


\ifextended
In the adopted model, each pollutant is defined by its mass $C(\vec{x},t)$ at a location $\vec{x}=(x,y,z)$ and time $t$. See Figure~\ref{fig:01}
for an illustration of the propagation of a pollutant, 
assuming a wind direction along the $x$ axis. The pollutant is emitted from the source at height $h$, and the concentration profiles are given in the downwind directions, using the dispersion factors $\sigma_z$ and $\sigma_y$.

\begin{figure*}[t!]
\centering
\includegraphics[width=0.65\textwidth]{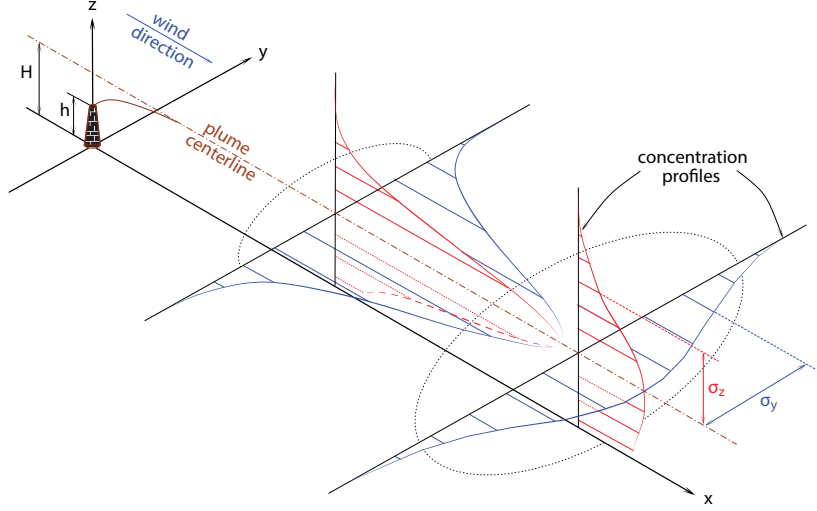}
\caption{An illustration of the Gaussian Plume model, adapted from J. M. Stockie
	\cite{stockie2011mathematics}.}
\label{fig:01}
\end{figure*}

Following~\cite{stockie2011mathematics}, the law of conservation of mass is  
\begin{equation}
\frac{\partial C}{\partial t}+\nabla\cdot\vec{J}=S\,,
\end{equation}
where $S(\vec{x},t)$ is a source or sink term, and $J(\vec{x},t)$ is the mass flux considering the effects of diffusion and advection. The equation reduces to
\begin{equation}
\frac{\partial C}{\partial t}+\nabla\cdot\left(C\vec{u}\right)=\nabla\cdot\left(\bf{K}\cdot\nabla C\right)+S\,,
\end{equation}
where $\vec{u}$ is the wind velocity and $\bf{K}$ is the $3\times 3$ diagonal matrix of the space-diffusion coefficients, which are assumed to be functions of the downwind distance only, and thus all equal to $K(x)$. After proceeding with the change of variable
\begin{equation}
r=\frac{1}{u}\int_{0}^{x}K(\xi)d\xi\,,\quad 
\end{equation} 

and assuming we either know the closed form of $K(x)$ from experimentally measured values or $K(x)$ is constant, 
the Gaussian Plume solution $c(r,y,z)$ of the advection-diffusion equation can then be derived, which for a homogeneous,
steady-state flow and a steady-state line source of finite length $L$ is given by: 
\begin{align}\label{plumelink}
\frac{Q_L}{2u\sqrt{\pi r}} \exp\!\left(-\frac{z^2}{4r}\right)\left[f\left(\frac{y+L/2}{2\sqrt{r}}\right)-f\left(\frac{y-L/2}{2\sqrt{r}}\right)\right]
,
\end{align}
where $Q_L$ is a emission constant rate, and where $f(x)=2/\sqrt{\pi}\int_{0}^x\exp(-\xi^2)d\xi$ is the error function. Note that, for low wind speeds, e.g. lower than $0.5~\si{m.s^{-1}}$, the diffusion term cannot be neglected in the $x$-direction relative to the advection term, rendering \eqref{plumelink} inaccurate. The interested reader may refer to~\cite{stockie2011mathematics}, page 359-360, for more details.
\fi

\subsection{The Implementation}

We used Caline \citep{caline}, the standard free 
dispersion-modelling suite, to solve the Gaussian Plume model for the inputs and outputs presented in the previous Section. We note while Caline
is one of the ``Preferred and Recommended Air Quality Dispersion Model''
of the Environmental Protection Agency in the USA~\citep{epa}, it has significant practical limitations. Specifically, it is limited to 20 line sources (of traffic) and 20 receptors (prediction points) per computational run, which in turn forces an arbitrary in-homogeneous discretisation of the road network and is a strong motivation for the use of our cross-domain deep learning framework. 

We implemented the approach for the use case of Dublin, Ireland. 
The area was partitioned into 12 domains, with 7-20 line sources of pollution in each subdomain. 
Inputs to the PDE solver comprised of hourly traffic volume data at each line source obtained (by aggregation of readings of traffic detectors) from the traffic control system (SCATS) used in Dublin, and hourly weather data (wind speed, direction, temperature, humidity) obtained from The Weather Company.
Available training data comprised
almost one year worth of hourly data
from July 1$^{\rm st}$ 2017 to May 2$^{\rm nd}$ 2018.
The outputs focused on concentrations of NO$_2$,  and  $\mathrm{PM}_{10}$ concentrations 
at predefined receptor locations, 
as illustrated in Figure~\ref{fig:Dublin}. To circumvent the limitation on number of receptors allowed by Caline (20 receptors per computational run), we ran the Caline model 15 times with receptors positioned in different locations to give adequate  spatial coverage of pollution estimates across the domain. 
Outputs were produced at 300 receptor locations within each subdomain giving a total of 3,600 estimates of pollution across the domain each hour for the 305-day study period.
Caline-model specific parameters were chosen for each subdomain $M_m$ based on the state-of-the-art practices~\citep{NAGENDRA20022083}: the emission factors based on the UK National Atmospheric Emissions Inventory database, dispersion coefficients based on the Caline recommendations (values for inner city, outer city areas), and background pollution levels chosen as the average time series values across the pollution measurements stations. 
This is in line with the usage of the Environmental Protection Agency of Ireland \cite{broderick2004validation}. {The background pollution levels were subtracted again from the Caline output to obtain an effective traffic contribution to the pollution concentrations. Each subdomain was then computed independently based on the specific traffic, weather, and model parameters for the locality.

\begin{figure*}[ht]
    \centering
    \includegraphics[width=0.45\textwidth]{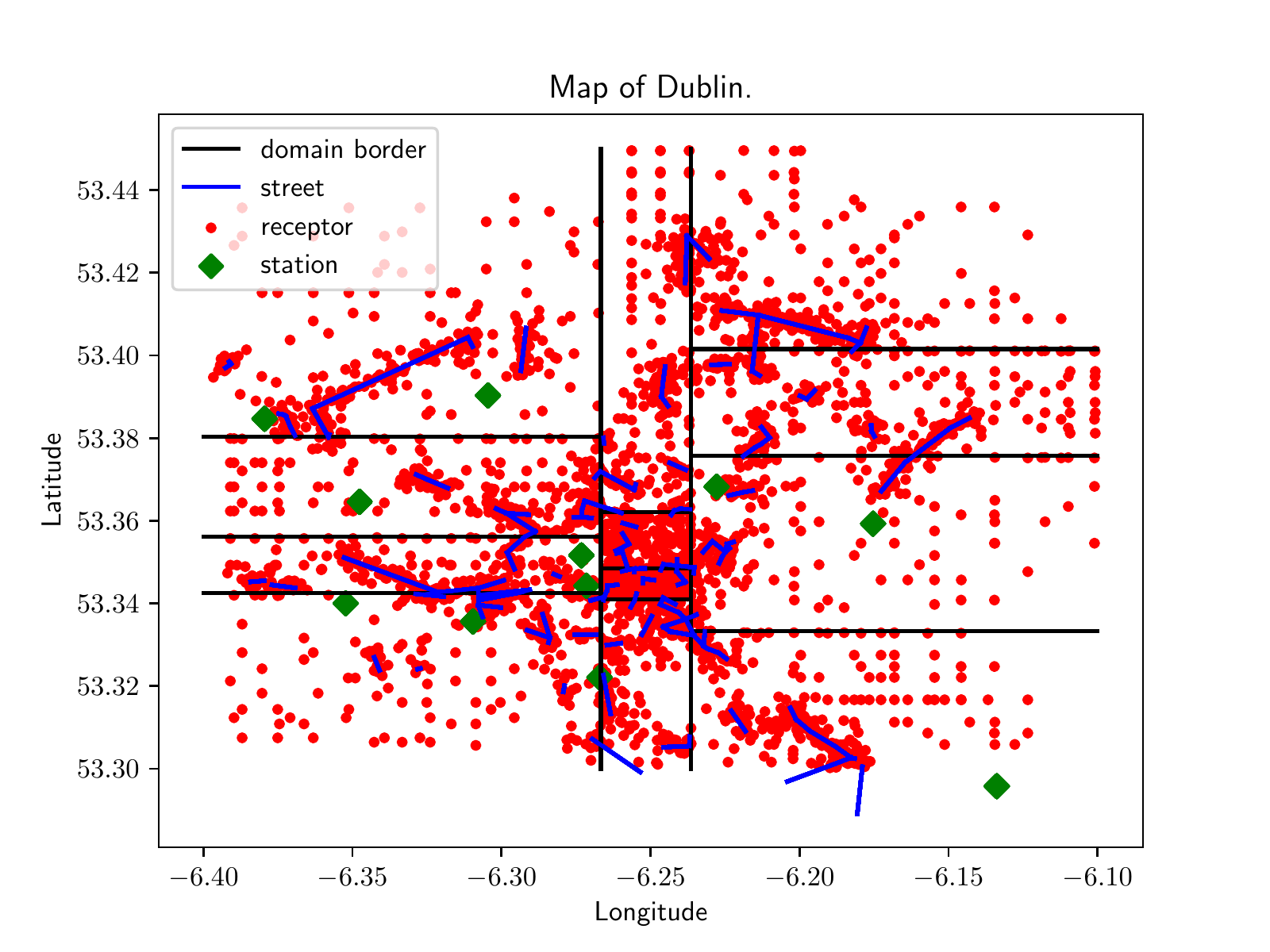}
    \includegraphics[width=0.45\textwidth]{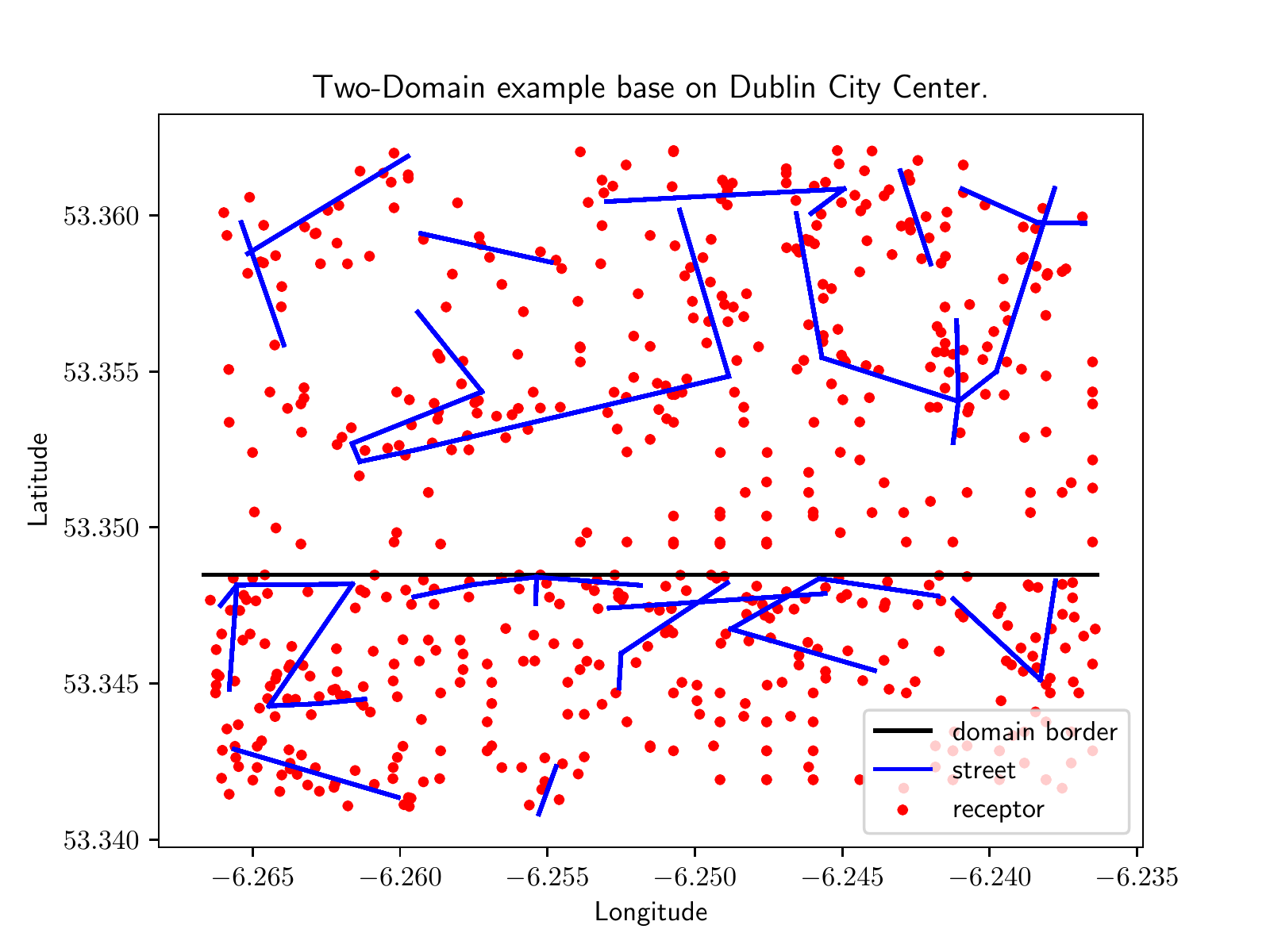} 
    \caption{\textbf{Left}: Map of Dublin, partitioned into 12 domains (black lines), displaying the positions of line sources (blue lines),  receptors (red dots), and measurement stations (green diamonds).
    \textbf{Right}: A close-up of Dublin city center, featuring two partitions with adjusted line sources close to the boundary (central partitions on the left).} 
    \label{fig:Dublin}
\end{figure*}

The RNN model was implemented in Tensorflow \citep{abadi2016} to obtain, in effect, the non-linear regression between the inputs and outputs, 
with the consistency constraints applied iteratively.
That is, with each map from the inputs to the outputs, we obtained further consistency constraints to use in subsequent runs on the same domain.
Features to the neural network consisted of the time step, the traffic line sources, the weather data, and a 
receptor location at each time step. Training label data consisted of the Caline outputs (without background pollution) for those features at the given receptor.
From those features, we created design matrices, each row consisting of the spatial coordinates of the start and end points of
 the traffic line sources (up to 40 coordinate tuples per subdomain) and traffic volume measurements for each of the line sources, 
 the weather data (wind speed, direction, directional standard deviation, temperature) for the locality, the coordinates of a receptor 
 for which the pollution concentrations should be estimated (there are 300 locations per subdomain), and the hourly timestamp measured in seconds since January 1\textsuperscript{st} 1970. 
The training data inherently has the structure of a time series, and as such it would be sensible to combine all receptor locations of a 
given subdomain in the input of that time slice, if our problem would be a temporal forecasting one. 
However, our problem of combining several subdomains by imposing consistency constraints across the boundaries is primarily a geospatial one.
For the consistency constraints, the models need to predict pollution concentrations at the subdomain boundaries, where no training data exists. That is: they need to learn the spatial relations of the underlying physics. 
As such, it is much more sensible to structure the learning set into spatial slices, predicting concentrations for one receptor at a time.

The timestamp input as well as the output labels were 
normalised between zero and one, whereas all other inputs were Gaussian normalised over each unit domain.
Overall, the feature design matrix, $\mathbf{X}$, had 1,494,900 rows (number of hourly estimates of all 300 receptors over the near-yearlong period) 
and up to 107 columns (number of features for each subdomain estimate). 
For subdomains with fewer line sources, $\mathbf{X}$ had correspondingly fewer columns. 
The label design matrix, $\mathbf{Y}$, for model training was composed of the 1,494,900 Caline model runs, 
each of which comprises one Caline estimate of pollution for each receptor location within each subdomain. 
This process was repeated for each of the 12 subdomains and for each of the two pollutants considered 
(NO$_2$ and $\mathrm{PM}_{10}$), resulting in a total of 24 corresponding $\mathbf{X}$ and $\mathbf{Y}$ 
matrices provided as an input to the RNN model.

The cost function used the standard regularisation with the $\ell_2$-loss of the weights. 
For the consistency constraints, we choose $\zeta = 1, \epsilon_{mn} = 1$, and considered different values for $\lambda$ and $\kappa$. The final topology consisted of a multilayer perceptron (MLP) with seven layers, each having 50 nodes, with a leaky ReLU activation function with $\alpha = 0.1$ after each layer \citep{goodfellow2016deep}.
Network training adopted the Adam optimisation algorithm with stochastic gradient descent on batches of size 128.
The $\mathbf{X}$ and $\mathbf{Y}$ data were always randomly shuffled into two groups to form the training-data set composed of 90\% of the 1,494,900 rows of data with the test-data set the remaining 10\%.
We trained for 25 epoch per iteration, with the consistency constraints between subdomains updated each iteration 
for a total of 20 iterations.

The complete source code has been released publicly under Apache license at \url{https://github.com/IBM/pde-deep-learning},
whereby further details of the implementation can be studied in detail. Likewise, all our data are freely available.
Traffic-detector data used in the research are freely available from Dublin City Council. 
Weather data have been obtained from The Weather Company, an IBM business, under a licence. 
While we do not have rights to redistribute The Weather Company data, 
a free API key can be obtained to download the data from the vendor.
Suggestions as to the model parameters are freely available from the Environmental Protection Agency, Ireland.
The Caline package used for comparison is freely available from the California Department of Transportation (Caltrans). 
Pollution measurement data used in the comparison are freely available from Dublin City Council 
and Environmental Protection Agency, Ireland.
We hope that this release of code and data stimulates further research in the field.

\begin{figure*}[t!]
    \centering
    \includegraphics[width=0.49\textwidth]{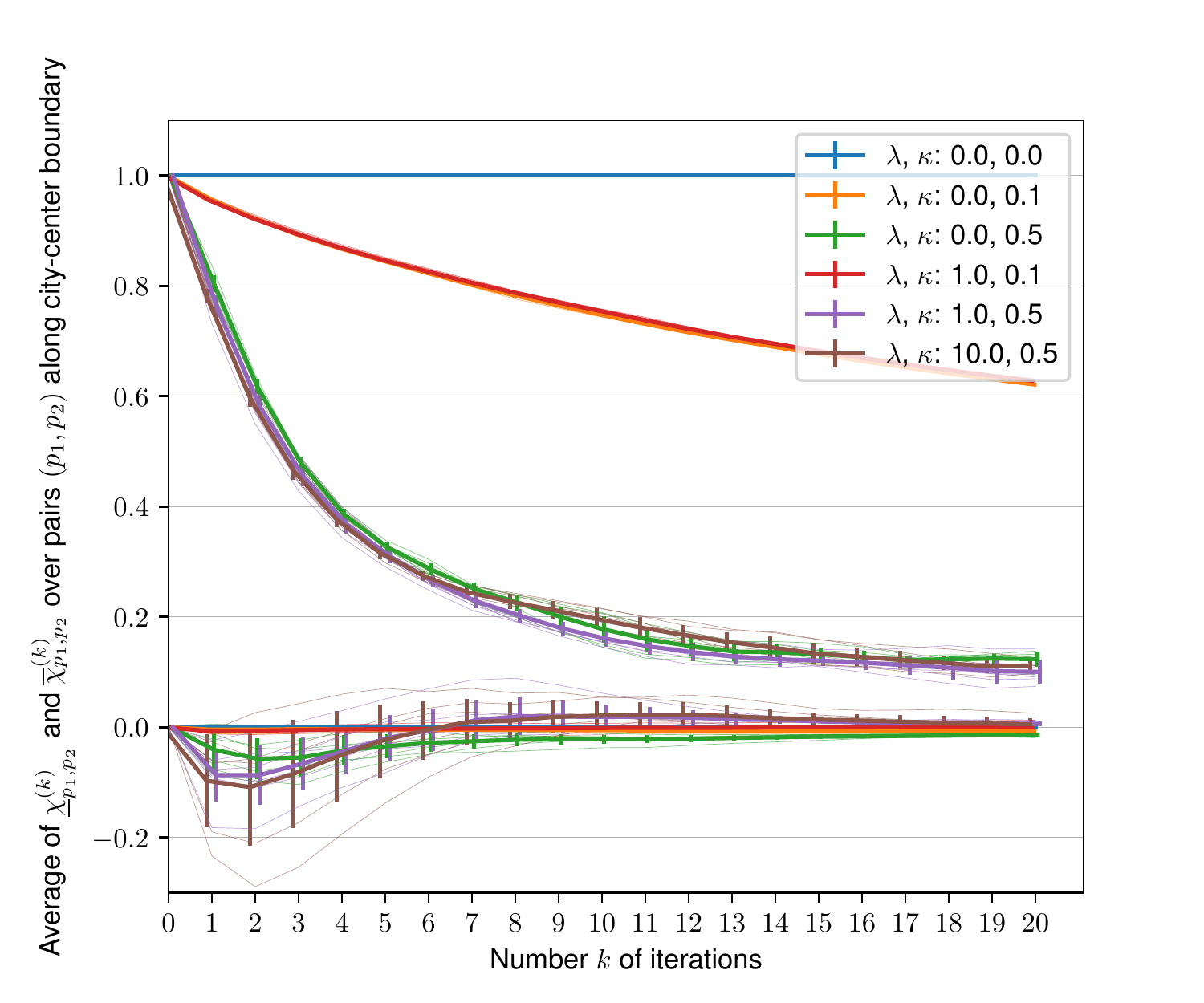}
    \includegraphics[width=0.49\textwidth]{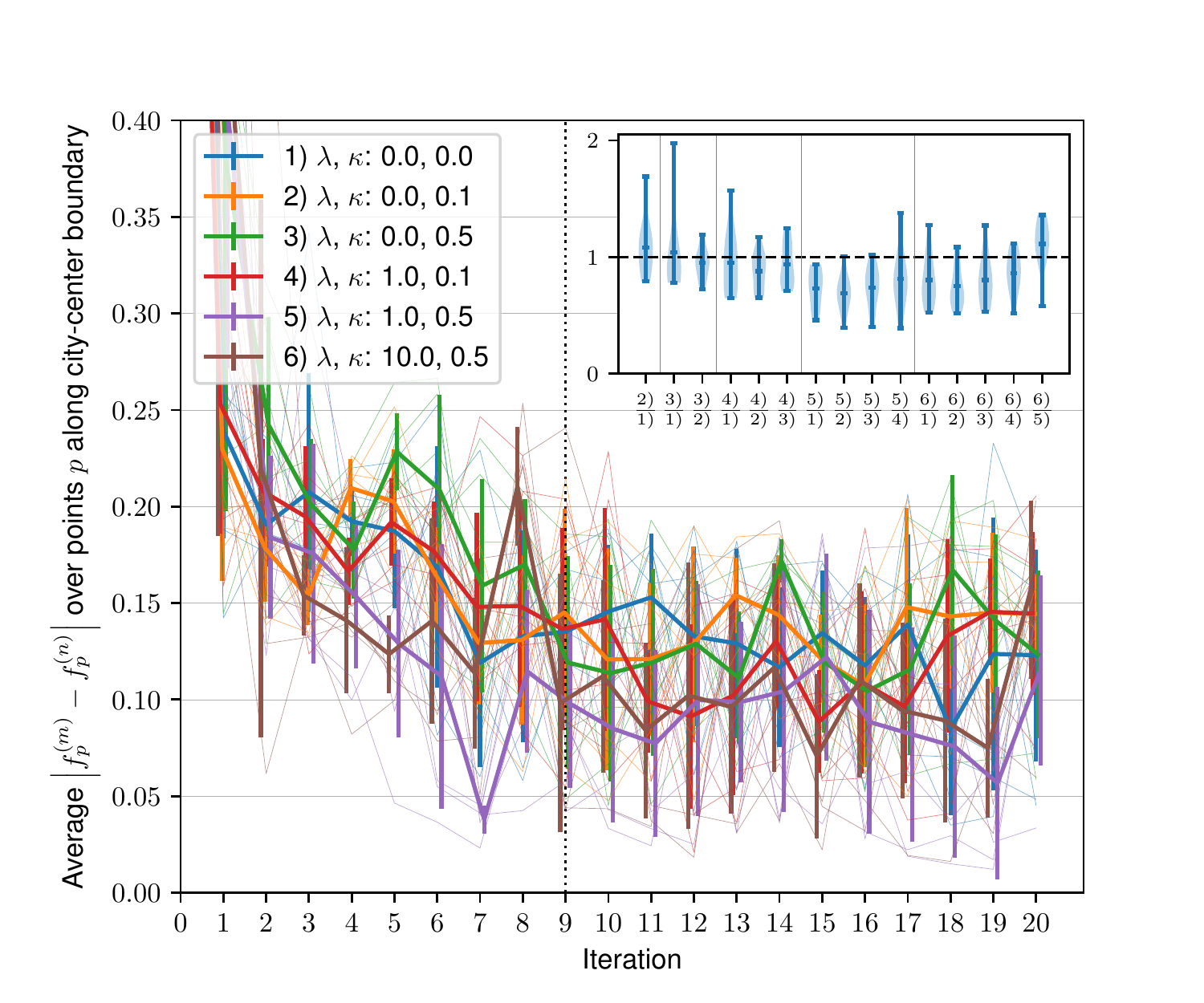}
    \caption{Effects of consistency constraints illustrated on two city-centre subdomains presented on the right in Figure~\ref{fig:Dublin}. The individual experiments are shown as thin lines. Their average is shown as a thick line with error bars at one standard deviation.
    \textbf{Left}: Mean lower ($\underline\chi$) and upper ($\overline\chi$) values of the consistency constraint interval plotted across the iterations, i.e., two lines per colour.
    \textbf{Right}: Mean difference between output of the DL models along the boundary between the two subdomains illustrating how specification of appropriate consistency constraints reduces the difference between predictions at adjacent boundaries. The inset shows the ratios between this value for the different parameter specifications. The ratio is computed over all iterations greater or equal than nine (dashed lined) to allow training error to stabilise.}
    \label{fig:chi}
\end{figure*}

\section{Results}
\label{sec:res}

First, to illustrate the workings of our approach, let us plot the convergence behaviour of the interval $(\underline \chi, \overline \chi)$ of \eqref{eq:updates} for the consistency constraints at the city-centre boundary for a few parameter combinations, as well as the convergence behaviour of the difference between the predicted values from the two neighbouring models along this boundary. In order to have clear boundary effects, we adjusted the street layout by bringing the streets in the southern subdomain closer to the boundary and pushing the streets in the northern subdomain further away, as illustrated on the right of Figure~\ref{fig:Dublin}.
In particular, Figure~\ref{fig:chi} (left) presents the evolution of the interval $(\underline \chi_{p_1, p_2}^{(k)},
\overline \chi_{p_1, p_2}^{(k)})$ over iterations $k$, when averaged over the pairs $(p_1, p_2)$ 
of corresponding points on the boundary of the two subdomains.
Clearly, we observe that $\underline \chi$ converges to $\overline \chi$%
, with rapid convergence especially in the first four iterations.
Further, one may add that faster convergence with increasing $\kappa$ is observable up to a certain point.
For higher values of $\kappa$, e.g., $\kappa = 0.7$, one enters an oscillatory regime (not shown),
which should be avoided.
This behaviour can be understood by drawing an analogy between \eqref{eq:updates} and accelerated first-order optimisation methods:
 one can think of $\kappa$ as a learning rate.
One should like to point out two caveats, though.
First, this behaviour is stochastic: Notice the difference between the individual sample paths, which is due 
to the non-convexity of the problem and the variable performance of randomised algorithms therein.
Second, this behaviour also does not translate to the values in $x^{(m)}, x^{(n)}$ being 
the same, except for $\lambda$ sufficiently large and rather impractical.
With these caveats, the behaviour demonstrates an iterative relaxation of the solution at neighbouring interfaces towards a reconciliation of both solutions.


The mean absolute error of the deep-learning model stabilises for all parameter sets after eight iterations. Figure~\ref{fig:chi} (right) demonstrates the convergence of the average difference of the predicted concentration values across the boundary. The inset shows the distributions of the ratios between the different parameter sets after eight iterations and demonstrates that imposing the consistency constraints does lead to a reduction of the discontinuity in predicted concentration values across the boundary of about $25\%$ to $30\%$ (c.f. ratios between $\lambda>0$ (in particular 5) and 6)\,) and $\lambda=0$ (\,1), 2), and 3)\,). The fact that all other ratios are close to one confirms that this effect is indeed systematic.

As a further point, let us mention the run-time of model training and model application. 
As has been stressed in the introduction, computational expense is one of the primary challenges of 
large-scale PDE-based models, limiting the geographical extent and resolution that can be studied. 
On the other hand, machine-learning approaches have a much lower computational expense at the model-application phase, i.e., once trained.
The computational expense of training the model, which 
	can be conceptually compared to the parameter estimation effort for PDE-based models, is non-negligible.
Considering commodity-compute resource (i.e., 2.3 GHz Intel Core i5 processor), 
training the entire domain for 20 iterations took about 120 CPU-hours.
(We note that this considers the use of the CPU only, and does not use general-purpose 
	graphics processing unit or other accelerators in the training or application of the RNN.
	Obviously, the wall-clock time can be significantly shorter with the use of dedicated GPU resources 
	and parallel computing. Also, four iterations $k = 4$ bring much of the improvement, as suggested in the previous paragraph.) 
This is obviously a significant 
	expense and one wishes to avoid frequent re-training of the model. 
We have trained on almost 
	a year of data to generate a robust model.
Retraining or updating of the model is not 
	anticipated, unless the model is to be applied to a different area or under markedly 
	different conditions (e.g., significant increase in the use of electric vehicles). 
In contrast, once trained, the computational cost of deploying the model to predict is negligible. 
Comparing the performance of the Caline model with the trained RNN model, we observed 
a speed-up factor of more than two orders of magnitude in the model application for the study period. 

\begin{figure*}[t!]
\centering
    \includegraphics[width=0.8\textwidth]{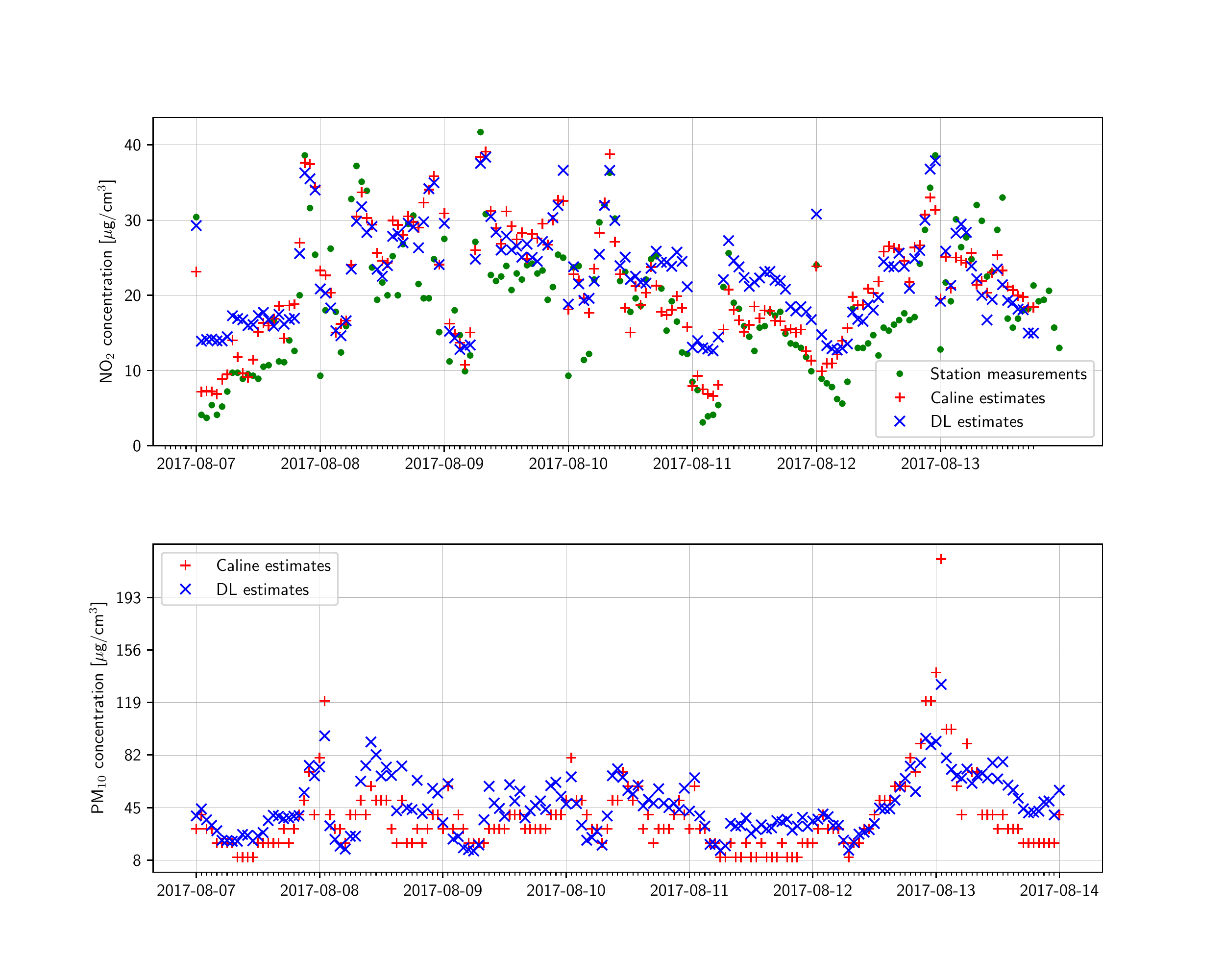}
    \caption{Time series of Caline (label data for the DL model) and DL estimates for one sample receptor over a one week time inverval. \textbf{Top}: NO$_2$ concentrations with added background values for unit-restored Caline and DL output plotted against measurements from a collocated measurement station over a 1-week time interval. \textbf{Bottom}: PM$_{10}$ concentrations for unit-restored Caline and DL output.}
    \label{fig:ts}
\end{figure*}

Last but not least, let us comment on the predictive skill of the DL model.
Performance evaluation considered the ability of the DL to replicate Caline estimates at 
defined locations with known traffic contributions to pollution, and more broadly across 
the entire city with highly-varying contributions of traffic to pollution.  Figure~\ref{fig:ts} presents a time-series plot of DL estimates against Caline for both NO$_2$ and 
PM$_{10}$ at one example receptor collocated with an NO$_2$ measurement site used for our validation. 
The neural-network closely captures the general trends of the Caline estimates, particularly the 
diurnal component, with (traffic-driven) higher values during the day.
Differences between Caline and DL model during that period are on average less than 3\,$\mu$g/cm$^3$ for NO$_2$ and 15\,$\mu$g/cm$^3$ for PM$_{10}$, with no evident biases.

\begin{figure*}[t!]
\centering
    \includegraphics[width=\textwidth]{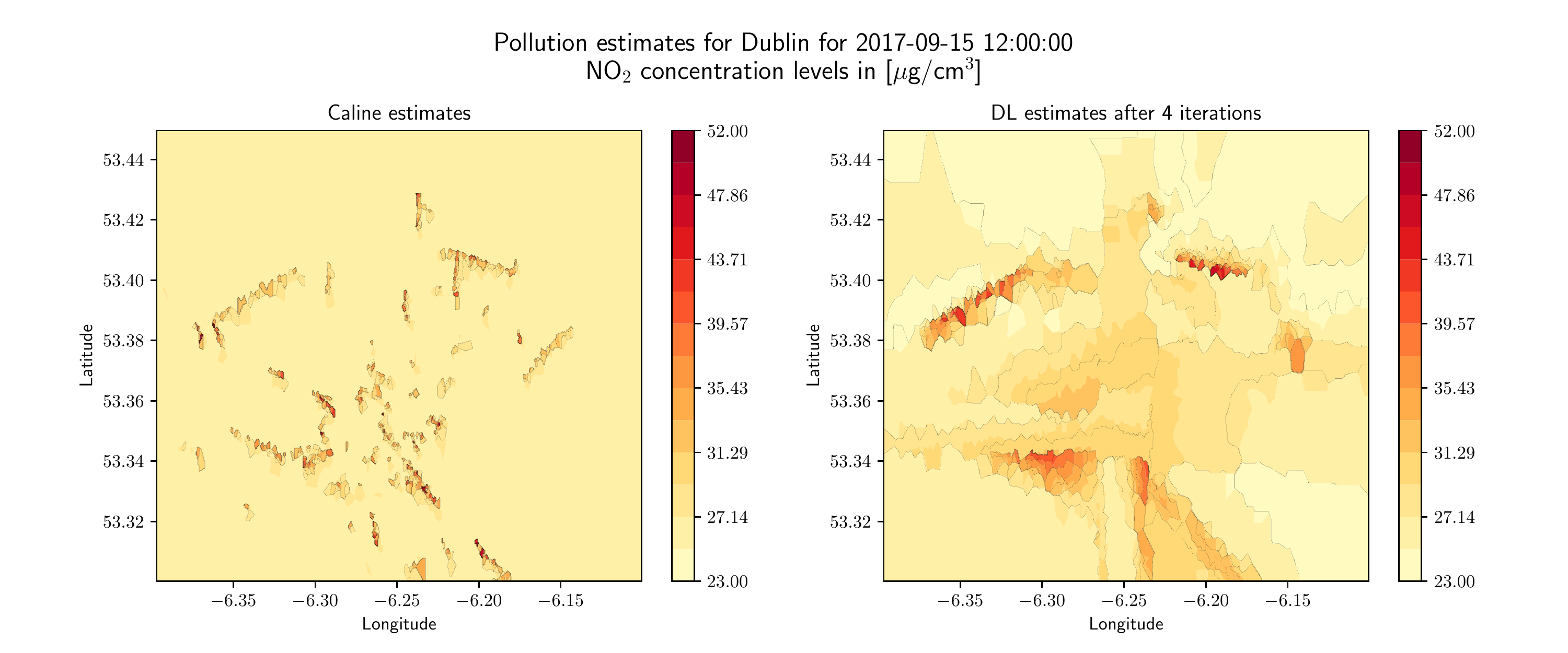}
    \caption{Spatial contour plot of modelled NO$_2$ estimates for selected time snapshot across the entire study domain. 
		The partitioning of the domain into individual subdomains and the distribution of line sources 
		of traffic volumes within those subdomains is provided in Figure \ref{fig:Dublin}.
	\textbf{Left}: Caline estimate. \textbf{Right}: Our DL estimate. }   
    \label{fig:contour}
\end{figure*}

The spatial distribution of pollutants is significantly more complex, encapsulating a high dependency 
on local features (location of traffic line sources) together with the Gaussian-plume distribution 
characteristics of the Caline model.
Figure~\ref{fig:contour} presents a contour map of Caline and DL estimates of NO$_2$ values across 
the entire domain at an instance in time.
Results demonstrate that the DL model captures areas of high pollution contributions rather well, 
with peak values similar for Caline and DL. 
Across large areas of the model domain, Caline reports low values of traffic pollution with values 
falling back to an ambient level of background pollution. 
The DL model, on the other hand, predicts a much more uniform distribution of NO$_2$ concentrations,
which is not as tightly restricted by the geographical proximity to traffic line sources. 

This serves to illustrate one of the key differences of model estimates guided by physical rules 
and that driven purely by data. Caline estimates of pollution are restricted to the immediate vicinity 
of line-sources based on the physical equations governing the Gaussian plume model.
The DL model faces no such restriction and instead seeks the optimal combinations of weights that minimise
 the objective functions. 
Results demonstrate a tendency towards a smoother distribution of pollution by the DL model compared to that produced by Caline.  
This results in a significant mismatch between DL and Caline estimates in regions where traffic-generated pollution is low.  
The mean absolute error (MAE) of the deep-learning computed values was 1.7\,$\mu$g/cm$^3$ 
with a standard deviation of 2.1\,$\mu$g/cm$^3$.


\section{Conclusions}
\label{sec:conclusions}




We have presented consistency constraints, which make it possible to train surrogate
models on small domains and apply the trained models to larger domains, while allowing 
incorporation of information external to the domain. 
The consistency constraints will ensure that the solutions are physically meaningful 
even at the boundary of the small domains in the output of the surrogate model. 
We have demonstrated promising results on an air-pollution forecasting model 
for Dublin, Ireland.

For the first time, this work makes it possible to apply deep-learning techniques to develop
\textit{surrogate models} that potentially exceed the capabilities of the more complex parent model.
Borrowing domain-decomposition techniques from the PDE community, it provides a framework
to merge the outputs from disparate models or solutions that have spatial dependencies. 
In contrast, traditional machine-learning approaches consider each model prediction to be a function of events that
happen within the computational domain. 
Numerous approaches, however, exist in the PDE literature
to incorporate processes outside the domain (e.g., Dirichlet or Neumann boundary conditions,
Iterative Schwarz interface methods) and are particularly common in parallel-computing implementations.
This paper presents a first demonstration of implementing exchange of information
across domains to a deep-learning framework. Leveraging consistency constraints, we demonstrate
a deep-learning approach that learns the mapping of each domain individually and using RNN techniques,
iteratively adjusts consistency constraints to provide an optimal representation of the \textit{global} solution.
Within the domain of our example application, recent surveys \citep{bellinger2017systematic} also suggest that ours is the 
first use of deep-learning in the forecast of air-pollution levels.

This work makes possible numerous extensions. 
Following the copious literature on PDE-based models of air pollution, one could consider further pollutants 
such as ground-level ozone concentrations~\citep{mallet2013minimax}, and ensemble \cite{odonncha2019ensemble} or 
 multi-fidelity methods.
One may also consider a joint model, allowing for traffic forecasting, weather forecasting, and air-pollution 
forecasting, within the same network, possibly using LSTM units 
\cite{cui2018high}, at the same time. 
More generally, one could consider further applications 
of the consistency constraints, e.g., in energy conservation, or merging the outputs of a 
number of PDE models within multi-physics applications.
In multi-resolution approaches, 
lower-resolution (e.g., city-, country-scale) component could constrain higher-resolution
components (e.g., district-, city-scale), which in turn impose consistency constraints on the former. 
In some applications, it may be useful to explore other network topologies. Following \cite{wiewel2018latent}, one could use
long short-term memory (LSTM) units.
Further, over-fitting control could be 
based on an improved stacked auto-encoder architecture~\citep{zhou2017delta}.
In interpretation of the trained model, the approach of~\cite{cui2018high}
may be applicable.
One could also consider applications to inverse problems, following \cite{RAISSI2017683,RAISSI2018125}.
Finally, 
one could 
generalise our methods in a number of directions of the 
multi-fidelity  \citep{peherstorfer2018survey} modelling, e.g.,
 by combining the reduced-order and full-order models using adaptation, fusion, or filtering.

Our work could also be seen as an example of 
 Geometric Deep Learning   \citep{7974879},
 especially in conjunction with the use of mesh-free methods \citep{sirignano2017dgm},
such as the 3D point clouds \citep{qi2017pointnet},
 non-uniform meshing,
 or non-uniform choice of receptors within the meshes.
Especially for applications, where the grids are 
 in 3D or higher dimensions, the need for such techniques is clear.
More generally, one could explore links to isogeometric analysis of \cite{cottrell2009isogeometric},
which integrates solving PDEs with geometric modelling.
Overall, the scaling up of deep learning for PDE-based models
seems to be a particular fruitful area for further research.

\section*{Acknowledgements}
\noindent
This work was in part supported by the European Union Horizon 2020 Programme (Horizon2020/2014-2020), under grant agreement no. 68838. The the first two authors would like to thank the Institute of Pure and Applied Mathematics, UCLA, for their hospitality during one part of the project.


\bibliographystyle{elsarticle-num-names}
\bibliography{DL4PDE,refs,references,ref}

\end{document}